\documentclass{article}
\usepackage{times}
\usepackage{epsfig}
\usepackage{graphicx}
\usepackage{amsmath}
\usepackage{amssymb}
\usepackage{bbm, dsfont}
\usepackage{color}
\usepackage{cite}
\usepackage[numbers,sort]{natbib}

\usepackage{graphicx}
\usepackage{comment}
\usepackage[margin=1.5in]{geometry}
\usepackage[utf8]{inputenc} 
\usepackage{url}            
\usepackage{booktabs}       
\usepackage{amsfonts}       
\usepackage{color}
\usepackage{authblk}

\usepackage{amsmath}
\usepackage{etoolbox}
\usepackage{bm}
\usepackage{amsthm}
\usepackage{mathtools}
\usepackage{algorithm}
\usepackage{algpseudocode}
\usepackage{xspace}

\newcommand{\MyMapTemplateNoPrefix}[3]{\expandafter#1\csname#3\endcsname{#2{#3}}}
\forcsvlist{\MyMapTemplateNoPrefix {\def} {\mathbf} } {0,1,a,b,c,d,e, f, g, h, i, j, k, l, m, n, o, p, q, r, u, v, w, x, y, z} 

\newcommand{\MyMapTemplatePrefix}[4]{\expandafter#1\csname#3#4\endcsname{#2{#4}}} 
\forcsvlist{\MyMapTemplatePrefix {\def} {\mathcal}{c}} {A,B,C,D,E,F,G,H,I,J,K,L,M,N,O,P,Q,R,S,T,U,V,W,X,Y,Z}  
\forcsvlist{\MyMapTemplatePrefix {\def} {\mathbf} {b}} {A,B,C,D,E,F,G,H,I,J,K,L,M,N,O,P,Q,R,S,T,U,V,W,X,Y,Z} 
\forcsvlist{\MyMapTemplatePrefix {\DeclareMathOperator} {} {} } {trace, argmax, argmin, CE, BCE, Ent, MMD}

\def\ie{\emph{i.e.}\@\xspace}
\def\eg{\emph{e.g.}\@\xspace}
\def\etc{\emph{etc.}\@\xspace}
\title{Open-Ended Visual Question Answering by Multi-Modal Domain Adaptation}
\author[1]{Yiming Xu\thanks{This work was done during Yiming's internship at Futurewei Technologies.}}
\author[2]{Lin Chen}
\author[2]{Zhongwei Cheng}
\author[3]{Lixin Duan}
\author[4]{Jiebo Luo}
\affil[1]{Northwestern University}
\affil[2]{Futurewei Technologies}
\affil[3]{University of Electronic Science and Technology of China}
\affil[4]{University of Rochester}

\begin{document}

\maketitle

\begin{abstract}
We study the problem of visual question answering (VQA) in images by exploiting supervised domain adaptation, where there is a large amount of labeled data in the source domain but only limited labeled data in the target domain with the goal to train a good target model. A straightforward solution is to fine-tune a pre-trained source model by using those limited labeled target data, but it usually cannot work well due to the considerable difference between the data distributions of the source and target domains. Moreover, the availability of multiple modalities (\ie, images, questions and answers) in VQA poses further challenges to model the transferability between those different modalities. In this paper, we tackle the above issues by proposing a novel supervised multi-modal domain adaptation method for VQA to learn joint feature embeddings across different domains and modalities. Specifically, we align the data distributions of the source and target domains by considering all modalities together as well as separately for each individual modality.
Based on the extensive experiments on the benchmark VQA 2.0 and VizWiz datasets for the realistic open-ended VQA task, we demonstrate that our proposed method outperforms the existing state-of-the-art approaches in this challenging domain adaptation setting for VQA.

\end{abstract}

\section{Introduction}
The task of visual question answering (VQA) is  building a model to answer questions given an image-question pair. Recently, it has received much attention of the researchers in the area of computer vision~\cite{Zhou2015SimpleBF,strong_baseline_vqa,bottomup,ban,counting,pythia3}. VQA requires techniques from both image recognition and natural language processing, and most existing works use Convolutional Neural Networks (CNNs) to extract visual features from images and Recurrent Neural Network (RNNs) to generate textual features from questions, and combine them to generate the final answers. 

However, most of the existing VQA datasets are artificially created and thus may not be suitable as training data for real-world applications. For example, VQA 2.0~\cite{vqa2.0} and Visual7W~\cite{visual7w}, arguably two of the most popular datasets for VQA, were created using images from MSCOCO~\cite{MSCOCO} with questions asked by crowd workers. Therefore, the images are typically of high quality and the questions are less conversational than the reality. On the contrary, the recently proposed VizWiz~\cite{vwz} dataset was collected from blind people who take photos and ask questions about them. Therefore, the images in VizWiz are often of poor quality, and questions are more conversational while some of the questions might be unanswerable due to the poor quality of the images. The VizWiz dataset reflects  more realistic setting for VQA, but its size is much smaller due to the difficulty of collecting such data. A straightforward method to solve this problem is to first train a model on the VQA 2.0 dataset and then fine-tune it using the VizWiz data. This solution can only provide limited improvement. There are two major issues. First, the VQA datasets are constructed in a different way, making them differ significantly in visual features, textual features and answers. \cite{cross_dataset} did an experiment to classify different VQA datasets with a simple multi-layer perceptron (MLP) of one hidden layer, which achieved over 98\% accuracy. This is a strong indication of the significant bias across different datasets. Our experiments also show that directly fine-tuning the model trained on VQA 2.0  results in minor improvement on VizWiz. Second, the two modalities (visual and textual) also pose a big challenge on the generalizability across datasets. It is challenging to consistently bridge the domain gap in a coordinated fashion when multiple modalities are involved due to the nature of the multi-modal heterogeneity with no common feature representations.

Domain adaptation methods, which handle the difference between two domains, have been developed to address the first issue~\cite{DA1,DA2,adda,dann,wdgrl,semida1,semida2,semida3}. However, most of the existing domain adaptation methods focus on single-modal tasks such as image classification and sentiment classification, and thus may not be directly applicable to multi-modal settings. On the other hand, these methods usually are subject to a strong assumption on the label distribution in that the source domain and the target domain share the same (usually small) label space, which may be  unrealistic in real-world applications. \cite{mmacm} proposed a new framework for unsupervised multi-modal domain adaptation, but it did not target at the VQA tasks. Recently, several VQA domain adaptation methods have been proposed to address the multi-modal challenge. However, to the best of our knowledge, all the existing VQA domain adaptation methods focus on the multiple choice setting, where several answer candidates are provided and the model only needs to select one from them. 
In contrast, we focus on a more challenging open-ended setting where there is no prior knowledge of answer choices and the model can select any term from a vocabulary.

In this paper, we address the aforementioned challenges by proposing a novel multi-modal domain adaptation framework.  We develop a method under the framework which can simultaneously learn  a domain invariant and downstream-task-discriminative multi-modal feature embedding based on an adversarial loss and a classification loss. We additionally incorporate the maximum mean distance (MMD) to further reduce the domain distribution mismatch
for multiple modalities, \ie,  visual embeddings, textual embeddings and joint embeddings. We conduct experiments on two popular VQA benchmark datasets. The results show that the proposed model outperforms the state-of-the-art VQA models and the proposed domain adaptation method surpasses other state-of-the-art domain adaptation methods on the VQA task. Our contributions are summarized as follows:
\begin{itemize}
    \item We propose a novel supervised multi-modal domain adaptation framework.
    
    \item We tackle the more challenging open-ended VQA task with the proposed domain adaptation method. To the best of our knowledge, this is the first attempt of using domain adaptation for open-ended VQA.
    
    \item The proposed method can simultaneously learn domain invariant and downstream-task-discriminative multi-modal feature embedding with an adversarial loss and a classification loss. At the same time, it minimizes the difference of cross-domain feature embeddings jointly over multiple modalities.
    
    \item We conduct extensive experiments between two popular VQA benchmark datasets, VQA 2.0 and VizWiz, and the results show the proposed method outperforms the existing state-of-the-art methods.
\end{itemize}

\section{Related Works}\label{sec:related_work}

\noindent\textbf{VQA Datasets} Over the past few years, several VQA datasets~\cite{visual7w,vqa2.0,vwz,visualgenome,vqa1.0} and tasks were proposed to encourage researchers to develop algorithms that answer visual questions. One limitation of many existing datasets is that they were created either automatically or from an existing large vision dataset like MSCOCO~\cite{MSCOCO}, and the questions were either generated automatically or contrived by human annotators on Amazon Mechanical Turk (AMT). Therefore, the images in these datasets are typically of high quality but the questions are less conversational. They might not be directly applicable to real-world applications such as~\cite{vwz} which aims to answer the visual questions asked by blind people in their daily life. The main differences between \cite{vwz} and other artificial VQA datasets are as follows: 1) Both the image and question quality of \cite{vwz} are lower as they suffer from poor lighting, out of focus and audio recording problems like clipping a question at either end or catching background audio content; 2) The questions can be unanswerable since blind people cannot verify whether the images contain the visual content they are asking about, due to blurring, inadequate lighting, framing errors, finger covering the lens,~\etc Our experiments also reveal that fine-tuning the model trained on the somewhat artificial VQA 2.0 dataset provides limited improvement on VizWiz, due to the  significant difference in bias between these two datasets.

\noindent\textbf{VQA Settings}\quad There are two main VQA settings, namely multiple choice and open-ended. Under the multiple choice setting, the model is provided with multiple candidates of answers and is expected to select the correct one from them. VQA models following this setting usually take characteristics of all answer candidates like word embeddings as the input to make a selection~\cite{cross_dataset,multiplechoicevqa}. However, in the open-ended setting, there is neither prior knowledge nor answer candidates provided, and the model can respond with any free-form answers. This makes this setting more challenging and realistic~\cite{ban,strong_baseline_vqa,pythia3,bottomup}.


\noindent\textbf{VQA Models}\quad Recently, a plethora of VQA models were proposed by researchers~\cite{Zhou2015SimpleBF,strong_baseline_vqa,bottomup,ban,pythia3}. Most of them consist of image and question encoders, and a multi-modal fusion module followed by a classification module. \cite{strong_baseline_vqa} used an  LSTM to encode the question and a residual network~\cite{resnet} to compute the image features with a soft attention mechanism. \cite{bottomup} implemented a bottom-up attention using Faster R-CNN~\cite{fasterrcnn} to extract features of detected image regions, and then a top-down mechanism used task-specific context to predict an attention distribution over the image regions. The final output was generated by an MLP after fusing the image and question features. \cite{ban} used a bilinear attention between two groups of input channels on top of low-rank bilinear pooling which extracted the joint representations for each pair of channels. \cite{pythia3} proposed an approach that takes original image features, bottom-up attention features from object detection module, question features and the optical character recognition (OCR) strings detected from the image as the input, and answers either with an answer from the fixed answer vocabulary or by selecting one of the OCR strings detected in the image. Similar to the state-of-the-art model~\cite{pythia3}, our VQA base model also takes original image features, bottom-up attention features and question features to predict the final answer. Details of our VQA base model is described in the next section.

\noindent\textbf{Domain Adaptation}\quad Domain adaptation techniques have been proposed to learn a common domain invariant latent feature space where the distributions of two domains are aligned. Recent works typically focused on transferring neural networks from a labeled source domain to a target domain where there is no or limited labeled data~\cite{DA1,DA2,adda,wdgrl,dann,semida1,semida2}. \cite{DA1} optimized for domain invariance to facilitate domain transfer and used a soft label distribution matching loss to transfer information between tasks. \cite{adda} proposed a framework which combines discriminative modeling, untied weight sharing and a GAN loss to reduce the difference between domains. \cite{wdgrl} estimated empirical Wasserstein distance between the source and the target samples and optimized the feature extractor network to minimize the estimated Wasserstein distance in an adversarial manner. \cite{dann} utilized gradient reversal layer to incorporate the training process of domain classifier, label classifier and feature extractor to align domains. Similarly, \cite{semida2} simultaneously minimized the classification error, preserved the structure within and across domains, and restricted similarity on target samples. The major difference between our work and these works is that we propose a novel multi-modal domain adaptation framework, while these works assumed a single modality.

\noindent\textbf{Domain Adaptation for VQA}\quad Although domain adaptation has been successfully applied to computer vision tasks, its applicability to VQA has yet to be well-studied. There was a recent work that investigated domain adaptation for VQA~\cite{cross_dataset}. It reduces the difference in statistical distributions by transforming the feature representation of the data in the target domain. However, one major limitation is the assumption of a multiple choice setting, where four answer candidates are provided as the input to the model. It is unrealistic in real-world applications because one can never guarantee that the ground truth answer is among four candidates.  Moreover, it is unclear how to create answer candidates for an image-question pair. On the contrary, our model is only provided with an image-question pair and can generate any free-form answers. This makes our task more challenging and realistic.




\section{The VQA Framework}

\begin{figure*}[h]
\centering
\includegraphics[width=0.98\linewidth]{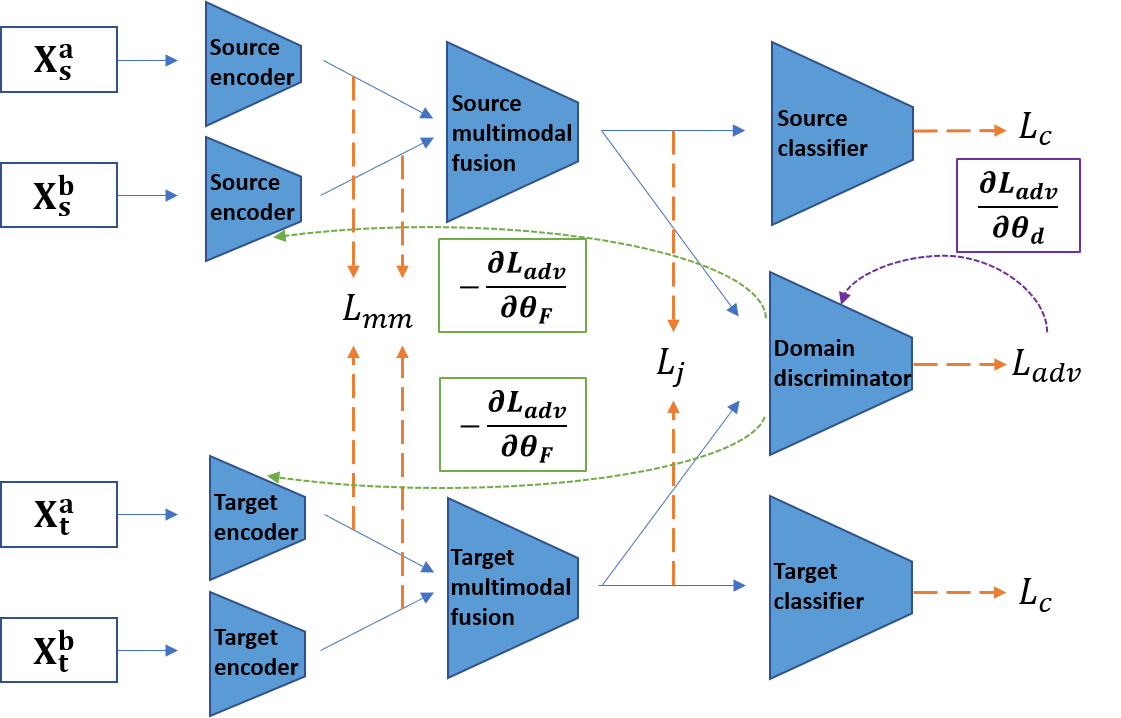} 
\caption{The proposed multi-modal domain adaptation framework. $\bX_s^a, \bX_s^b, \bX_t^a, \bX_t^b$ denote original features for two modalities. The blue arrow denotes forward propagation while the orange arrow denotes the loss calculation. The purple and green arrows denote backward propagation for discriminator loss $L_{adv}$. Note that the sign is reversed when the loss backpropagates through the gradient reversal layer in domain discriminator.}\label{fig:framework}
\label{fig2}
\end{figure*}


In this section, we describe our base VQA framework. Given an image $I$ and a question $Q$, the VQA model estimates the most likely answer $\hat{a}$ from a large vocabulary based on the content of the image, which  can be written as follows:
\begin{equation}
    \hat{a} = \argmax_a P(a|I,Q).
\end{equation}
Our base framework consists of four components: 1) a question encoder; 2) an image encoder; 3) a multi-modal fusion module; and 4) a classification module at the output end. We will elaborate each component in the following subsections.

\noindent{\bf Question Encoding}\quad The question $Q$ of length $T$ is first tokenized and encoded using  word embedding based on pretrained GloVe~\cite{glove} as $S=\{\x_0,\x_1,...,\x_T\}$. These embeddings are then fed into a GRU cell~\cite{gru}. The encoded question is obtained from the last hidden state at time step $T$ denoted as $\q=f^q(Q; \bm{\theta}_q) \in \cR^{d_q}$, where $f^q(Q; \bm{\theta}_q)=\h_T$, $\h_t= \text{GRU}(\x_t, \h_{t-1}; \bm\theta_q)$ for $1\leq t\leq T$, and $d_q$ is the feature dimension.

\noindent{\bf Image Encoding}\quad Similar to \cite{bottomup} and \cite{pythia3}, we first feed the input image $I$ to an object detector~\cite{Detectron} pretrained on the Visual Genome dataset \cite{visualgenome} based on Feature Pyramid Networks (FPN)~\cite{fpn} with ResNeXt~\cite{resnext} as the backbone. The output from the fully connected $fc6$ layer is used as the region-based features, \ie, $\bV_r=\{\v_1, \v_2,..., \v_K\}$ with $\v_i$ as the feature for $i$-th object. In the meanwhile, we divide the entire image into a $7\times7$ grid, and obtain the grid-based features $\bV_g$ by average pooling features from the penultimate layer $5c$ of a pretrained ResNet-101 network~\cite{resnet} on ImageNet dataset. Finally, we combine $\bV_r$ and $\bV_g$ as well as question embedding $\q$ to obtain the joint feature embedding in a multi-modal fusion module as described in the next paragraph.

\noindent{\bf Multi-Modal Fusion and Classification}\quad
The question embedding $\q$ is used to obtain the top-down,~\ie~region-based attention on image features $\bV_r$. Then, the region-based features $\bV_r$ are averaged based on the attention weights to obtain the weighted region-based image features. Similarly, grid-based features $\bV_g$ are fused with question embedding $\q$ by concatenation. The fused grid-based features and the weighted region-based image features are then concatenated to obtain the final image features $\v$. We denote the final image feature embedding as $\v=f^v(\q,I;\bm\theta_v)$. The final joint embedding $\e=f^j(\q,\v)$ is then calculated by taking the Hadamard product of $\q$ and $\v$, which is then fed to an MLP $f^c(\e;\bm\theta_c)$ for classification, \ie, $a = f^c(\e; \bm\theta_c)$. The final answer is represented by $\hat{a} = \argmax_{a} f^c(\e; \bm\theta_c)$.

\section{Multi-Modal Domain Adaptation}
In this section, we present our framework for supervised multi-modal domain adaptation. We assume there are two modalities\footnote{For simplicity, we assume the data has two modalities, but it can be easily generalized to more modalities.} of source samples $\bX_s=[\bX^a_s, \bX^b_s]$, where $a$, $b$ denote the two modalities, and labels $\bY_s$ drawn from a source domain joint distribution $P_s(x, y)$, as well as the two modalities of target samples $\bX_t=[\bX^a_t, \bX^b_t]$ and labels $\bY_t$ drawn from a target joint distribution $P_t(x, y)$. We also assume there are sufficient source data so that a good pretrained source model can be built but the amount of target data is limited so that learning on only the target data leads to poor performance. Our goal is to learn target representations for two modalities $f^a_t$, $f^b_t$, multi-modal fusion $f^j_t$ and target classifier $f^c_t$ with the help of pretrained source representations $f^a_s$, $f^b_s$, $f^j_s$ and source classifier $f^c_s$. For the VQA task in our work, $a,b$ denote visual and textual modalities, respectively.

A typical approach to achieving this goal is to regularize the learning of the source and target joint representations by minimizing the distance of empirical distributions between the source and target domains, \ie, between $f^j_s \left(f^a_s(\bX^a_s;\bm\theta^a_s), f^b_s(\bX^b_s;\bm\theta^b_s);\bm\theta^j_s\right)$ and $f^j_t\left(f^a_t(\bX^a_t;\bm\theta^a_t), f^b_t(\bX^b_t;\bm\theta^b_t);\bm\theta^j_t\right)$. 
In this way, the data from the source domain and the target domain are projected onto a similar latent space, such that well-performing source model can lead to well-performing target model. Following this idea, we propose a novel multi-modal domain adaptation framework as shown in Figure~\ref{fig:framework}.

\subsection{Joint Embedding Alignment}
We propose to reduce the difference of joint embeddings between the source and the target domains by minimizing the Maximum Mean Discrepancy (MMD). The intuition is that two distributions are identical if and only if all of their moments coincide. Suppose we have two distributions $P_s$, $P_t$ over a set $\mathcal{X}$. Let $\varphi:\mathcal{X} \rightarrow \mathcal{H}$, where $\mathcal{H}$ is a reproducing kernel Hilbert space (RKHS). Then, we have:
\begin{flalign}
&\MMD(P_s, P_t) \nonumber\\
   &=\underset{\varphi \in \mathcal{H}, ||\varphi||_\mathcal{H}\leq 1}{\sup}\Big| \mathbbm{E}_{\bX_s\sim P_s} [\varphi(\bX_s)] - \mathbbm{E}_{\bX_t\sim P_t} [\varphi(\bX_t)] \Big| \nonumber\\
    &=||\mu_{P_s}-\mu_{P_t}||_\mathcal{H},
\end{flalign}
where $\mu_P=\int k(x,\cdot)P(dx)$ is the kernel mean embedding of $P$ and $k$ is a kernel function such as a Gaussian kernel. Let $\bX_s=\{\x^s_1,...,\x^s_{n_s}\}\sim P_s$ and $\bX_t=\{\x^t_1,...,\x^t_{n_t}\}\sim P_t$, the empirical estimate of the distance between $P_s$ and $P_t$ is  
\begin{eqnarray}
    \MMD(\bX_s, \bX_t)=\left\|\frac{1}{n_s}\sum_{i=1}^{n_s}\varphi(\x^s_i)-\frac{1}{n_t}\sum_{i=1}^{n_t}\varphi(\x^t_i)\right\|_\mathcal{H}.
\end{eqnarray} 
We then define the loss function as 
\begin{eqnarray}
    L_j=E_{\bX_s\sim p_s, \bX_t\sim p_t} \left[\MMD^2(\e_s, \e_t)\right], \label{eq:loss_Lj}
\end{eqnarray}
where 
$\e_s=f^j_s\left(f^a_s \left(\bX^a_s;\bm\theta^a_s\right), f^b_s(\bX^b_s;\bm\theta^b_s);\bm\theta^j_s\right)$ and
$\e_t=f^j_t\left(f^a_t(\bX^a_t;\bm\theta^a_t), f^b_t(\bX^b_t;\bm\theta^b_t);\bm\theta^j_t\right)$.
By minimizing the difference between source and target joint embeddings, we enforce that the joint embeddings of both source domain and target domain will be projected onto a similar latent space. 

\subsection{Multi-Modal Embedding Alignment}
It is more challenging to reduce multi-modal domain shift than conventional single-modal domain shift. The previous loss $L_j$ in Eq.~(\ref{eq:loss_Lj}) does not explicitly consider the multi-modal property. Aligning only the joint feature embedding is insufficient to adapt the source domain to the target domain. This is because the feature extractor for each modality has its own complexity of domain shift, which often differs from each other (\eg,~visual vs. textual). Aligning only the fused features cannot fully reduce domain differences. 

Therefore, we introduce the following term to minimize the maximum mean discrepancy  between every single modality, \ie, $\MMD\left(f^a_s(\bX^a_s;\bm\theta^a_s),f^a_t(\bX^a_t;\bm\theta^a_t)\right)$ 
and
$\MMD\left(f^b_s(\bX^b_s;\bm\theta^b_s),f^b_t(\bX^b_t;\bm\theta^b_t)\right)$.
Then, the loss function we try to minimize can be written as
\begin{eqnarray}
L_{mm} &\!\!\!\!\!\!=\!\!\!\!\!\!& E_{\bX_s\sim p_s, \bX_t\sim p_t} \left[\gamma_a \MMD^2\left(f^a_s(\bX^a_s; \bm\theta^a_s), f^a_t(\bX^a_t; \bm\theta^a_t)\right) \right. \nonumber\\
&\!\!\!\!\!\!+\!\!\!\!\!\!& \left. \gamma_b \MMD^2\left(f^b_s(\bX^b_s; \bm\theta^b_s), f^b_t(\bX^b_t; \bm\theta^b_t)\right)\right],
\end{eqnarray}
where $\gamma_a$ and $\gamma_b$ are trade-off parameters for two modalities.

\begin{figure*}[!ht]
\centering
\includegraphics[width=0.99\linewidth]{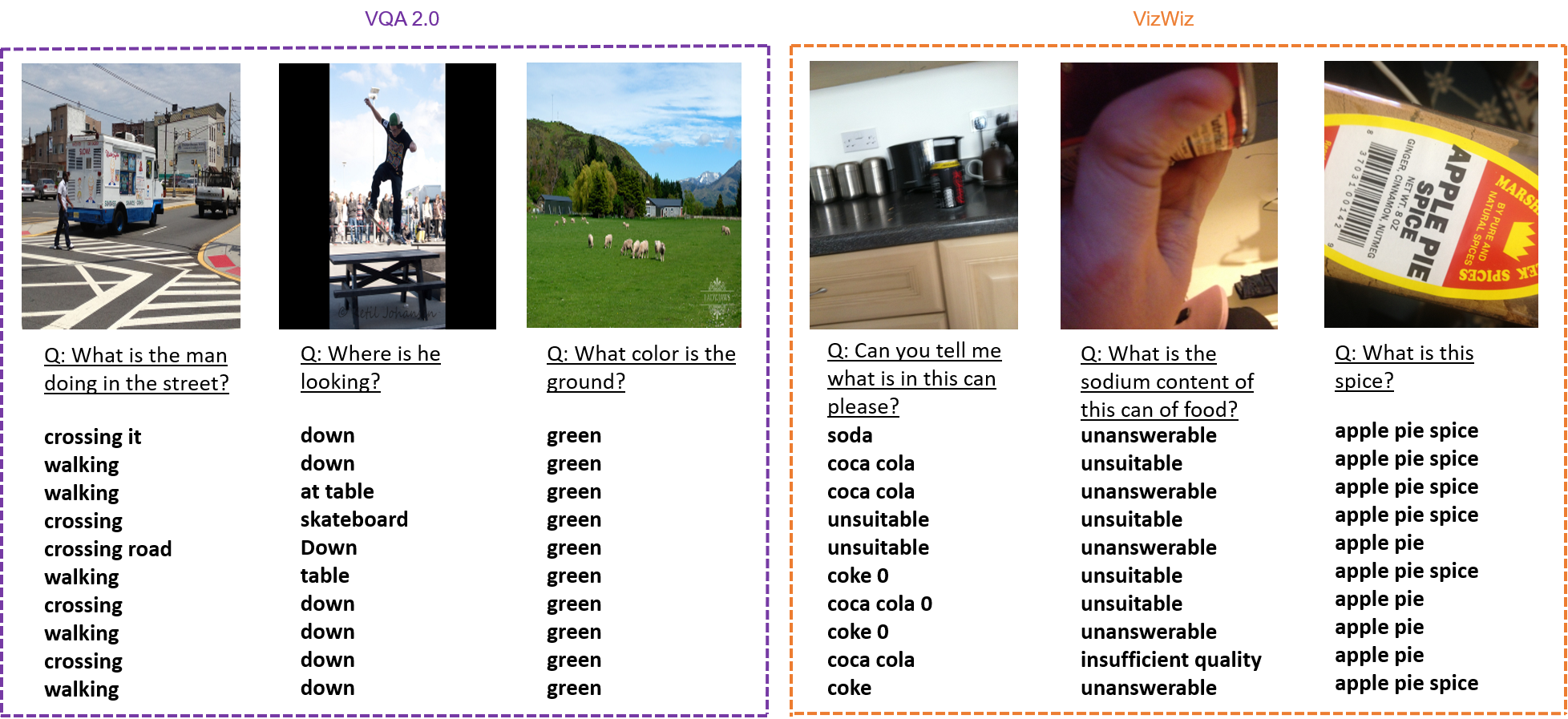}
\caption{Sample image-question pairs and valid answers for VQA 2.0 and VizWiz datasets. Please note that for each image-question pair, there are 10 answers provided by 10 different crowd workers.} 
\label{fig1}
\end{figure*}

\subsection{Classification}
While minimizing the distance between source and target embeddings, we also want to maintain the classification performance on both the source domain and the target domain. Similarly as in a standard supervised learning setting, we employ the cross entropy loss for classification:
\begin{eqnarray}
L_c=&E_{(\bX_t, \bY_t)\sim p_t} \left[CE(f^c_t(\e_t;\bm\theta^c_t),\bY_t)\right] \nonumber\\
&+ \gamma_c E_{(\bX_s,\bY_s)\sim p_s} \left[CE(f^c_s(\e_s;\bm\theta^c_s),\bY_s)\right],
\end{eqnarray}
where $CE$ denotes the standard cross entropy loss, and $\gamma_c$ is a trade-off parameter between the two domains.

\subsection{Domain Discriminator}
We also propose to use a domain classifier $f^d$ to reduce the mismatch between the source domain and target domain by confusing the domain classifier such that it cannot correctly distinguish a sample from source domain or target domain. The domain classifier $f^d$ has a similar structure to $f^c_t$ or $f^c_s$ except the last layer outputs a scalar in $[0, 1]$ with the value indicating how likely the sample comes from the source domain. Thus, $f^d$ can be optimized according to a standard cross-entropy loss. To make the features domain-invariant, 
the source and target mappings are optimized according to a constrained adversarial objective. The domain classifier minimizes this objective while the encoding model maximizes this objective. The generic formulation for domain adversarial technique is: 
\begin{eqnarray}
L_{adv}&=&-E_{\bX_s\sim p_s}\left[\log f^d(\e_s;\bm\theta_d)\right] \nonumber\\
&&- E_{\bX_t\sim p_t}\left[\log (1-f^d(\e_t;\bm\theta_d))\right].
\end{eqnarray}
For simplicity, we denote $\bm\theta^F = \left(\bm\theta^a_s, \bm\theta^a_t, \bm\theta^b_s, \bm\theta^b_t, \bm\theta^j_s, \bm\theta^j_t\right)$ as the parameters of all feature mappings and $\bm\theta^C=(\bm\theta^c_s, \bm\theta^c_t)$ as the parameters of all label predictors. Putting all together, we obtain our final objective function to minimize as follows:
\begin{eqnarray}
L(\bm\theta^F, \!\bm\theta^C, \!\bm\theta^d)\!=\! L_c + \lambda_j L_j + \lambda_{mm} L_{mm} - \lambda_{adv} L_{adv},
\end{eqnarray}
where we seek the parameters which attain a saddle point $\hat{\bm\theta}^F, \hat{\bm\theta}^C, \hat{\bm\theta}^d$ of $L$, satisfying the following conditions:
\begin{eqnarray}
\left(\hat{\bm\theta}^F, \hat{\bm\theta}^C\right)&=&\argmin_{\bm\theta^F,\bm\theta^C} L(\bm\theta^F, \bm\theta^C, \hat{\bm\theta}^d)\nonumber\\
\hat{\bm\theta}^d&=&\argmax_{\bm\theta^d} L(\hat{\bm\theta}^F, \hat{\bm\theta}^C, \bm\theta^d).
\end{eqnarray}


At the saddle point, the parameters $\bm\theta^d$ of the domain classifier minimize the domain classification loss $\cL_{adv}$ while the parameters $\bm\theta^C$ of the label predictor minimize the label prediction loss $\cL_c$. The feature mapping parameters $\bm\theta^F$ minimize the label prediction loss such that the features are discriminative, while maximizing the domain classification loss such that the features are domain-invariant. 

\section{Experiments}

In this section, we validate our method on the open-ended VQA task and compare it with state-of-the-art methods.

\subsection{Datasets}
Two standard VQA benchmarks are used in our experiments, VQA 2.0~\cite{vqa2.0} and VizWiz \cite{vwz}. A comparison of the statistics for these datasets are listed in Table~\ref{tab:dataset}, which shows that the scale of VizWiz is much smaller in terms of the numbers of images and questions. Although VizWiz has more unique answers, only 824 out of its top 3,000 answers overlap with the top 3,000 answers in VQA 2.0. This explains why models trained on VQA 2.0 perform poorly on VizWiz, and their limited transferability. We find 28.63\% of questions in VizWiz are even not answerable due to reasons mentioned before, making the domain gap even more significant. Figure~\ref{fig1} shows some examples from both VQA 2.0 and VizWiz datasets.

\begin{table}[t!]
\centering

\caption{The statistics of VQA 2.0 and VizWiz dataset. Numbers denote train/validation/test information, respectively.}

  \begin{tabular}{l|c|c }
  \hline
      & \textbf{VQA 2.0}&\textbf{VizWiz}\\ \hline
\# images  & 83K/41K & 20K/3K/8K\\
\# questions & 443K/214K/448K & 20K/3K/8K \\ 
\# answers & 4.4M/2.1M/NA & 0.2M/0.03M/NA \\ 
\# unique answers & 3,126 & 58,789\\\hline
  \end{tabular}
  \label{tab:dataset}
\end{table}

\subsection{Evaluation Metrics}
In VQA, each question is usually associated with 10 valid answers from 10 annotators. We follow the conventional evaluation metric on the open-ended VQA setting to compute the accuracy using the following formula:
\begin{eqnarray} 
Acc(a)=\frac{1}{K}\sum_{k=1}^K \min\left( \frac{\sum_{1\leq j\leq K, j\neq K} \mathbbm{1}(a=a_j) }{3}, 1\right),
\end{eqnarray} 
An answer is considered correct if at least three annotators agree on the answer. Note that the true answers in VizWiz test set are not publicly available.  In order to obtain the performance on the test set, results need to be uploaded to the official online submission system (\url{https://evalai.cloudcv.org/web/challenges/challenge-page/102}). 

\subsection{Implementation Details}
In all our experiments, we extract $K=100$ objects for each image to construct our region-based features $\bV_r$ and set the visual feature dimension to $2048$. We also set the hidden dimension of GRU to $1024$ and hidden dimension after fusion to $4096$. The question length is truncated at $24$. In the training phase, we apply a warm-up strategy by gradually increasing the learning rate $\eta$ from $0.001$ to $0.01$ in the first $2000$ iterations. It is then multiplied by $0.15$ after every $4000$ iterations. We use a batch size of $128$. 

For domain adaptation, we let the source and target networks share the same parameters up to the penultimate layer, \ie,~$\bm\theta^v=\bm\theta^v_s=\bm\theta^v_t$ and $\bm\theta^q=\bm\theta^q_s=\bm\theta^q_t$. In multi- or single-modal alignment, we use Gaussian kernel $k(x,y)=\exp^{-\frac{||x-y||^2}{2\sigma^2}}$ to compute MMD. The trade-off parameters are set as $\lambda_j=0.025$, $\lambda_{mm}=0.008$, $\gamma_v=0.8$, $\gamma_q=1$, $\gamma_c=0.001$, and $\lambda_{adv}=0.003$. 

\subsection{Experimental Setup}
First, we conduct experiments by using the VQA 2.0 dataset as the source domain and the VizWiz dataset as the target domain, to evaluate the effectiveness of our proposed method for multi-modal domain adaptation. We also conduct experiments in the opposite way,  using VizWiz as the source domain and VQA 2.0 as the target domain, to further demonstrate the effectiveness of our approach.

We need to emphasize that we choose not to use an overly  strong base model (\ie,~question embedding from FastText, complex fusion techniques, OCR tokens~\etc), as our focus is on multi-modal adaptation instead of the base model itself. Despite that, we will show that our proposed domain adaptation method with a weaker base model still outperforms the fine-tuned state-of-the-art model.

\subsection{Results and Analysis}
\noindent\textbf{Adaptation from VQA 2.0 to VizWiz}\quad As discussed in previous sections, we first pretrain a source model on the VQA 2.0 dataset, and then adapt the pretrained source model to the target dataset VizWiz. The results of our proposed method and other leading methods are shown in Table~\ref{tab:vqa_vwz}.

We first compare our method with the original VizWiz baseline proposed by \cite{vwz}, the previous state-of-the-art VQA model BAN by \cite{ban} and the current state-of-the-art VQA model Pythia by \cite{pythia3}. It is clear that our method outperforms the state-of-the-art models by a significant margin from Table~\ref{tab:vqa_vwz}.

In order to validate that the better performance of our method is not due to a strong base model, we additionally report the results of our method in Table~\ref{tab:vqa_vwz_three}, with 1) training our single base model from scratch using only the VizWiz dataset ({\bf Target only}), 2) fine-tuning from the model pretrained on the VQA 2.0 dataset ({\bf Fine-tune}), and 3) our proposed domain adaptation method ({\bf DA}). From Table~\ref{tab:vqa_vwz_three}, it shows that our model fine-tuned from VQA 2.0 is about $0.75$ percent worse than Pythia fine-tuned from VQA 2.0 ($53.97\%$ vs. $54.72\%$), indicating that the better performance of our final model than the state-of-the-art is not from a strong base model. Moreover, the accuracy of our base model trained from scratch is $53.11\%$, falling behind $0.6$ percent to Pythia trained from scratch, which is consistent with our observation that our method even with a weaker base model can achieve superior final results.

\begin{table}[!t]
\centering
\caption{Accuracy (in \%) of different methods on VizWiz.} 
  \begin{tabular}{l  | c }
  \hline
      \textbf{Method} & \quad\textbf{Accuracy}\\ \hline
VizWiz baseline  &\quad47.50\\\hline
BAN  &\quad51.40\\\hline
Pythia\footnote{Please note that, the $54.72\%$ accuracy for Pythia was obtained by fine-tuning from the model pretrained on the VQA 2.0 dataset.} & \quad54.72\\ \hline
Ours &\quad\textbf{55.87} \\ \hline
  \end{tabular}
  \label{tab:vqa_vwz}
\end{table}

\begin{table}[t]
\centering
\caption{Accuracy (in \%) comparison for our base model. {\bf Target only} denotes training from scratch, {\bf Fine-tune} means fine-tuning and {\bf DA} presents our domain adaptation method.} 
  \begin{tabular}{c|c|c}
  \hline\textbf{Target only}&\quad\textbf{Fine-tune}&\quad\textbf{DA}\\ \hline
53.11&\quad53.97&\quad55.87 \\ \hline
  \end{tabular}
  \label{tab:vqa_vwz_three}
\end{table}

\begin{table*}[h!]
\centering
\caption{Results breakdown into different categories of different methods for domain adaptation from VQA 2.0 to VizWiz. Breakdown numbers are performance on VizWiz \textit{test-dev} split.} 
  \begin{tabular}{l | c | c |c|c|c}
  \hline
     (Accuracy in \%) & \quad\textbf{Overall}&\quad\textbf{Yes/No}&\quad\textbf{Number}&\quad\textbf{Answerable}&\quad\textbf{Other}\\ \hline
VizWiz baseline  &\quad47.50&\quad66.90&\quad22.00&\quad77.00&\quad29.40\\\hline
BAN  &\quad51.40&\quad68.10&\quad17.90&\quad\textbf{85.30}&\quad31.50\\\hline
Pythia &\quad54.22&\quad\textbf{74.83}&\quad31.11&\quad84.08&\quad35.03 \\ \hline
Ours&\quad\textbf{55.87}&\quad74.33&\quad\textbf{32.00}&\quad83.32&\quad\textbf{38.53} \\ \hline
  \end{tabular}
  \label{tab:vqa_vwz_breakdown}
\end{table*}


\noindent\textbf{Results breakdown into answer categories}\quad Table \ref{tab:vqa_vwz_breakdown} shows the accuracy breakdown into different answer categories. The results show that our model achieves new state-of-the-art performance on ``Number'' and ``Other'' categories as well as overall accuracy. Note that the overall accuracy for Pythia in this table is $54.22\%$ instead of $54.72\%$ which we were unable to reproduce using the released code and there are no breakdown numbers reported associated with it. The best we can achieve with Pythia (after fine-tuning from VQA 2.0) is $54.22\%$ and the corresponding breakdown numbers are reported in the table.

\begin{table}[t!]
\centering
\caption{Ablation study of our proposed method.} 
  \begin{tabular}{l| c | c }
  \hline
      \textbf{Method}&\textbf{Accuracy}&\textbf{Improvement}\\ \hline
Target only  & 53.11& - \\\hline
(+ Fine-tune) & 53.97&+ 0.86 \\ \hline
+ MMD on V and Q, CLS& 55.46 &+ 1.49\\\hline
+ MMD, GRL on joint& 55.87 &+ 0.41\\\hline
+ Ensemble of 3 models& 56.20 &+ 0.33\\\hline
  \end{tabular}
  \label{tab:ablation}
\end{table}

\noindent\textbf{Ablation study}\quad  We conduct an ablation study to show the contributions of different components of our method. Specifically, we consider: \textbf{1. Target only:} Training the base model using only the data in the target domain. \textbf{2. +Fine-tune:} Pretrain a model on the source VQA 2.0 dataset and then fine-tune the model on the target VizWiz dataset. Please note that this is unavailable during adaptation thus it is marked inside ``()''. \textbf{3. +MMD on V and Q, CLS:} 
Our domain adaptation method with MMD alignment on visual and textual features separately, and classification modules applied for both domains.
\textbf{4. +MMD, GRL on joint:} 
Our domain adaptation method with MMD alignment also on the joint embeddings of both domains, along with the domain discriminator by gradident reversal layer. \textbf{5. +Ensemble of 3 models}. 
The results show that the multi-modal MMD brings the most significant performance gain, which validates that aligning on every single modality is beneficial to the transferability of multi-modal tasks. In addition, MMD on joint embedding and discriminator is also crucial to bring further performance gain. Not surprisingly, an ensemble of three models pushes our performance even higher to $56.20\%$, which is the state-of-the-art performance to date.


\noindent\textbf{Comparisons on domain adaptation methods}\quad We compare our multi-modal domain adaptation method with some popular domain adaptation methods, including DANN~\cite{dann}, ADDA~\cite{adda}, WDGRL~\cite{wdgrl}, and SDT~\cite{DA1}. Note that DANN, ADDA and WDGRL were originally designed for unsupervised domain adaptation. For fair comparison, we fine-tune the model using target labels after unsupervised adaptation (hence they are indicated by a suffix `+'). SDT is currently the most popular and best-performing supervised domain adaptation method. The results shown in Table~\ref{tab:DA_comparison} illustrate that compared to direct fine-tuning, the existing domain adaptation methods do not help much (DANN performs even worse) in the multi-modal task, while our method outperforms both direct fine-tuning and existing domain adaptation methods by a notable margin. 

\begin{table}[!h]
\centering
\caption{Accuracy (in \%) comparisons of our method with state-of-the-art domain adaptation methods. } 
  \begin{tabular}{l|c }
  \hline
      \textbf{VizWiz}&\quad\textbf{Accuracy}\\ \hline
Fine-tune  & 53.97 \\\hline
DANN+ & 53.65 \\ \hline
ADDA+ &54.06 \\ \hline
WDGRL+& 54.28 \\\hline
SDT&54.56\\\hline
Ours&\textbf{55.87}\\\hline
  \end{tabular}
  \label{tab:DA_comparison}
\end{table}

\begin{table}[t!]
\centering
\caption{Results comparison using less data.} 
  \begin{tabular}{l|c | c |c}
  \hline
       \textbf{Target data used}& \textbf{Target only }&\textbf{Fine-tune}&\textbf{DA}\\ \hline
1/8  & 39.51&43.39 & 45.02\\\hline
1/4  & 43.75&47.71 & 48.93\\\hline
1/2 & 47.48&50.12&52.32 \\ \hline
All data& 53.11&53.97&55.87 \\ \hline
  \end{tabular}
  \label{tab:fewer_data}
\end{table}

\noindent\textbf{Adaptation with fewer target training samples}\quad We also validate the robustness of our framework by reducing the target training dataset size. We experiment with various target sizes of 1/8 (2,500), 1/4 (5,000), 1/2 (10,000) and all data (20,000). The results are shown in Table~\ref{tab:fewer_data}. We can observe that with the increase of the amount of training data, the performance gain over fine-tuning is decreasing. We conjecture that this is because when we have limited amount of target data, having more prior knowledge is beneficial to model performance, while having more target data will make prior knowledge less helpful. However, our method can stably improve the performance because it sufficiently makes use of target data and source data. 
It is more promising that our domain adaptation method using fewer samples can achieve comparable or better performance compared to training from scratch using doubled amount of data (especially when target data is scarce), \eg, our method using 1/4 data (48.93\%) outperforms training from scratch using 1/2 data (47.48\%).

\begin{table}[!h]
\centering
\caption{Accuracy (in \%) comparison for our single base model adapted from VizWiz to VQA 2.0.} 
  \begin{tabular}{c|c|c}
  \hline
      \textbf{Target only} & \quad\textbf{Fine-tune}&\quad\textbf{DA}\\ \hline
 68.89&\quad69.25&\quad70.06 \\ \hline
  \end{tabular}
  \label{tab:vwz_vqa}
\end{table}

\noindent\textbf{Adaptation from VizWiz to VQA 2.0}\quad In order to further validate the robustness of our method, we reverse the source domain and the target domain and perform adaptation. We pretrain the source model on VizWiz and adapt the source model to VQA 2.0. The results are shown in Table \ref{tab:vwz_vqa}, from which we still can observe a significant improvement for our method against fine-tuning. As a comparison, the performance of BAN and Pythia trained from scratch are 69.08\% and 69.21\%, and our DA model achieves comparable performance to the state-of-the-art on VQA 2.0.

\section{Conclusion}


We have presented a novel supervised multi-modal domain adaptation framework for open-ended visual question answering. Under the framework,  we have developed a new method for VQA which can simultaneously learn domain-invariant and downstream-task-discriminative multi-modal feature embedding. We validate our proposed method on two popular VQA benchmark datasets, VQA 2.0 and VizWiz, in both directions of adaptation. The experimental results show our method outperforms the  state-of-the-art methods.

\bibliography{biblio}
\bibliographystyle{ieee}

\end{document}